\documentclass{article}

\usepackage{microtype}
\usepackage{graphicx}
\usepackage{subfigure}
\usepackage{booktabs} %

\usepackage{hyperref}

\usepackage[accepted]{icml2025}

\usepackage{amsmath}
\usepackage{amssymb}
\usepackage{mathtools}
\usepackage{amsthm}

\usepackage[capitalize,noabbrev]{cleveref}

\usepackage{multirow}

\theoremstyle{plain}

\theoremstyle{definition}

\theoremstyle{remark}

\usepackage[textsize=tiny]{todonotes}

\usepackage{colortbl,caption}

\definecolor{citecolor}{HTML}{0071BC}
\definecolor{linkcolor}{HTML}{ED1C24}
\definecolor{Gray}{gray}{0.86}
\newcolumntype{g}{>{\columncolor{Gray}}c}     %

\hypersetup{colorlinks=True, citecolor=citecolor, urlcolor=magenta}

\icmltitlerunning{Landsat-Bench}

\begin{document}

\twocolumn[
\icmltitle{Landsat-Bench: Datasets and Benchmarks for Landsat Foundation Models}

\begin{icmlauthorlist}
\icmlauthor{Isaac Corley}{wherobots}
\icmlauthor{Lakshay Sharma}{msft}
\icmlauthor{Ruth Crasto}{msft}
\end{icmlauthorlist}

\icmlaffiliation{wherobots}{Wherobots}
\icmlaffiliation{msft}{Microsoft}

\icmlcorrespondingauthor{Isaac Corley}{isaac@wherobots.com}

\icmlkeywords{Remote Sensing, Geospatial Foundation Models, Landsat}

{%
\vspace{0.75em}
\centering
\includegraphics[width=0.95\textwidth]{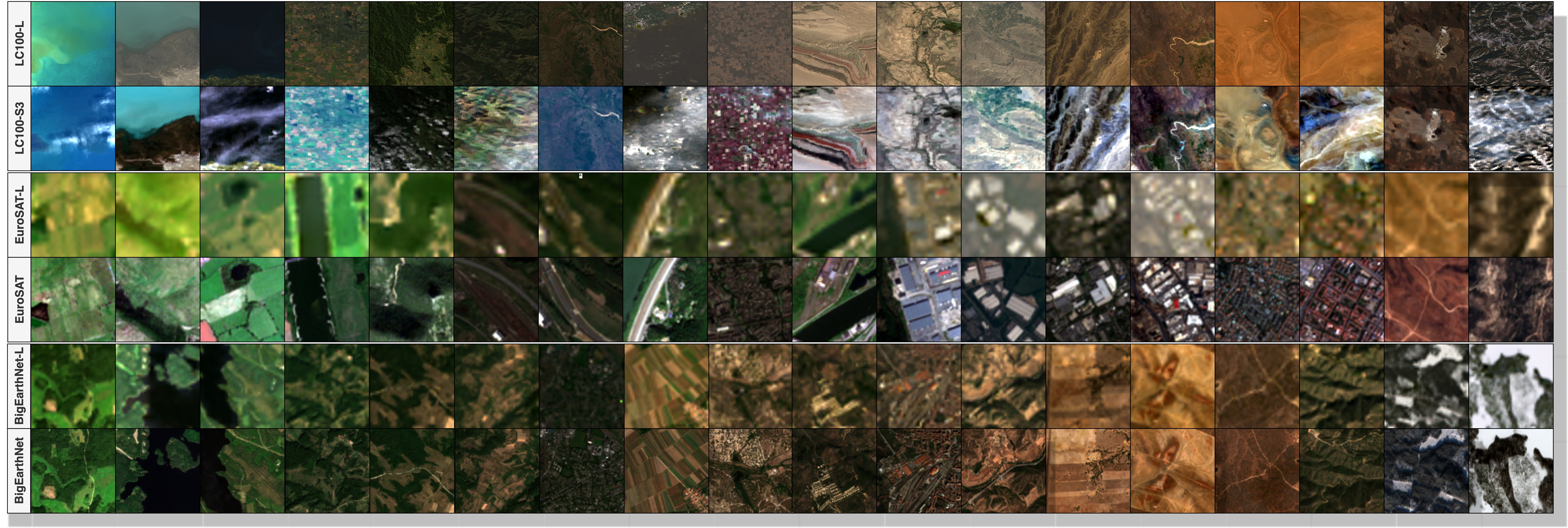}%
\captionof{figure}{\textbf{We introduce \texttt{Landsat-Bench}: a new benchmark for Geospatial Foundation Models (GFM) pretrained on Landsat imagery}. Our benchmark contains three Landsat variants (denoted with the suffix \textit{“—L”.}) of existing remote sensing datasets. Samples of each original dataset and our proposed Landsat variants are provided above (from top-to-bottom by row: \textit{LC100-L, LC100-S3, EuroSAT-L, EuroSAT, BigEarthNet-L \& BigEarthNet}).
}%
\label{fig:teaser}%
}

\vskip 0.3in
]

\printAffiliationsAndNotice{} %

\begin{abstract}
The Landsat program offers over 50 years of globally consistent Earth imagery. However, the lack of benchmarks for this data constrains progress towards Landsat-based Geospatial Foundation Models (GFM). In this paper, we introduce \texttt{Landsat-Bench}, a suite of three benchmarks with Landsat imagery that adapt from existing remote sensing datasets  -- EuroSAT-L, BigEarthNet-L, and LC100-L. We establish baseline and standardized evaluation methods across both common architectures and Landsat foundation models pretrained on the SSL4EO-L dataset. Notably, we provide evidence that SSL4EO-L pretrained GFMs extract better representations for downstream tasks in comparison to ImageNet, including performance gains of +4\% OA and +5.1\% mAP on EuroSAT-L and BigEarthNet-L. Code and datasets are available at \href{https://github.com/isaaccorley/landsat-bench}{github.com/isaaccorley/landsat-bench}.
\end{abstract}

\section{Introduction}
\label{sec;intro}
The Landsat program~\cite{roy2014landsat} has the longest-running continuous record of satellite-based Earth observation, capturing multispectral imagery of every continent with global, temporal, and spectral consistency in comparison to other openly available satellite constellations. This archive consists of petabytes of imagery across several decades, providing an opportunity for building Geospatial Foundation Models (GFM)~\cite{mai2023opportunities} with the potential to generalize across time, space, and environmental conditions. Unlike many modern satellite constellations, Landsat’s continuity across multiple generations (Landsat 1 to Landsat 9) makes it uniquely suited for modeling global-scale environmental dynamics, urbanization, and land use change.

\begin{table*}[t!]
\centering
\caption{\textbf{Comparisons of the datasets in \texttt{Landsat-Bench} and their original forms.} \textit{ML} and \textit{MT} indicate multi-label and multi-temporal datasets, respectively.}
\label{tab:datasets}
\resizebox{0.65\textwidth}{!}{%
\begin{tabular}{lcccccccc}
\toprule
\textbf{Dataset} &
\textbf{Sensor} &
\textbf{Samples} &
\textbf{Bands} &
\textbf{Res. (m)} &
\textbf{Image Size} &
\textbf{Classes} &
\textbf{ML} &
\textbf{MT} \\
\toprule

EuroSAT & Sentinel-2 & 27,000 & 13 & 10 & 64 × 64 & 10 & $\times$ & $\times$ \\
\rowcolor{Gray}
EuroSAT-L & Landsat 8 & 27,000 & 7 & 30 & 22 × 22 & 10 & $\times$ & $\times$ \\
\midrule

BigEarthNet & Sentinel-2  & 590,326 & 12 & 10 & 120 × 120 & 19 & \checkmark & $\times$ \\
\rowcolor{Gray}
BigEarthNet-L & Landsat 8&  590,326 & 7 & 30 & 40 × 40 & 19 & \checkmark & $\times$ \\
\midrule

LC100-S3 & Sentinel-3 & 21,692 & 21 & 300 & 96 × 96 & 23 & \checkmark & \checkmark \\
\rowcolor{Gray}
LC100-L & Landsat 8 & 21,690 & 7 & 30 & 960 × 960 & 23 & \checkmark & \checkmark \\

\bottomrule
\end{tabular}%
}
\end{table*}

Despite being publicly available, Landsat has been largely overlooked in the recent wave of GFM development, which has centered around higher-resolution or more frequently updated sensors like Sentinel-1, Sentinel-2, and NAIP~\cite{wang2025towards}. This neglect likely stems from a critical gap in benchmarking ability due to the scarcity of curated, standardized benchmarks for Landsat imagery. In this work, we aim to close this gap in favor of motivating more research into Landsat GFMs. With global consistency, decades of temporal depth, and well-studied downstream applications, Landsat has the ability to improve knowledge for representation learning in Earth observation if common benchmarks, e.g. Geo-bench~\cite{lacoste2023geo} and CopernicusBench~\cite{wang2025towards} are developed for researchers to empirically evaluate their progress{.

\paragraph{Motivation}
This work is largely inspired by the SSL4EO-L efforts of~\cite{stewart2023ssl4eo} to develop a large-scale pretraining dataset of Landsat imagery for building foundation models for use on downstream tasks. Due to the lack of available benchmark datasets for Landsat imagery, the authors were required to modernize two existing cloud detection datasets, L7 Irish~\cite{scaramuzza2011development} and L8 Biome~\cite{foga2017cloud}, and develop two semantic segmentation datasets by aligning Landsat imagery with pixel-level land cover annotations provided by the National Land Cover Database (NLCD)~\cite{dewitz2019nlcd} and Cropland Data Layer (CDL)~\cite{boryan2011monitoring}.

Prior to SSL4EO-L, SSL4EO-S12~\cite{wang2023ssl4eo} was developed as a large-scale dataset for pretraining self-supervised models on unlabeled satellite imagery from the Sentinel-1 SAR and Sentinel-2 multispectral (MSI) constellations. Due to the wide availability of benchmark datasets for the Sentinel program, the improvement of these foundation models was easily observed across several downstream tasks.

\paragraph{Related Work}
Utilizing existing dataset labels and georeferencing to align with other data products has become accessible due to the existence of catalogs and geospatial data querying tools such as Google Earth Engine (GEE)~\cite{gorelick2017google} and Microsoft's Planetary Computer~\cite{microsoft_open_source_2022_7261897}. \citet{wang2023feature} developed the EuroSAT-SAR dataset, a Sentinel-1 version of the EuroSAT benchmark dataset~\cite{helber2019eurosat}, using GEE to evaluate SAR foundation models. \citet{cong2022satmae} created a Sentinel-2 version of the fMoW dataset~\citep{christie2018functional} to evaluate their SatMAE architecture pretrained on Sentinel-2 imagery.

\paragraph{Contributions} In summary, there has recently been significant progress in developing large pretraining datasets for several satellite constellations; however, additional benchmark datasets for quantifying downstream task performance of satellite-specific foundation models are still needed. In this paper, we look to further close this gap. We detail our contributions below:

\begin{itemize}
\item We introduce the first iteration of \texttt{Landsat-Bench} consisting of three benchmark datasets for evaluating geospatial foundation models pretrained on Landsat imagery: \textbf{LC100-L}, \textbf{EuroSAT-L}, and \textbf{BigEarthNet-L}.

\item We provide baselines for our benchmarks of standard deep learning architectures and the SSL4EO-L line of geospatial foundation models.
\end{itemize}

\section{Datasets}
We introduce three new datasets for evaluating machine learning models trained on Landsat imagery -- \textit{LC100-L}, \textit{EuroSAT-L}, and \textit{BigEarthNet-L}. We use existing labeled remote sensing datasets commonly used in benchmarking foundation models trained on satellite imagery, namely the LC100-S3~\citep{wang2025towards}, EuroSAT~\citep{helber2019eurosat}, and BigEarthNet~\citep{sumbul2019bigearthnet} datasets. We utilize the original dataset implementations from the \texttt{torchgeo} library~\cite{stewart2022torchgeo}.  Descriptions of the datasets and our Landsat variants are provided in Table \ref{tab:datasets}. Below are details of each original dataset.

\begin{table*}[t!]
\centering
\caption{\textbf{Baseline evaluation of 9 pretrained models on \texttt{Landsat-Bench}}. We report both KNN (k=5) and Linear Probe (LP) results on extracted features of each method. For EuroSAT-L classification we report overall accuracy (OA). For BigEarthNet-L and LC100-L multilabel classification we report (micro) mean average precision (mAP). \textbf{*} denotes the model was pretrained on the SSL4EO-L dataset using the weights from \cite{stewart2023ssl4eo} made available in \texttt{torchgeo}.}
\label{tab:results}
\small                      %
\setlength{\tabcolsep}{5pt} %
\resizebox{0.98\textwidth}{!}{%
\begin{tabular}{@{}cccccccccccc@{}}
\toprule
 &
   & &
  \multicolumn{3}{c}{\textbf{ResNet-18 (11.7M)}} &
  \multicolumn{3}{c}{\textbf{ResNet-50 (25.6M)}} &
  \multicolumn{3}{c}{\textbf{ViT-S (22.1M)}} \\
  \cmidrule[0.75pt](lr){4-6}\cmidrule[0.75pt](lr){7-9}\cmidrule[0.75pt](lr){10-12}
\textbf{Method} &
  \textbf{Dataset} &
  \textbf{Zonal Stats} &
  \textbf{ImageNet} &
  \textbf{MoCo v2*} &
  \textbf{SimCLR*} &
  \textbf{ImageNet} &
  \textbf{MoCo v2*} &
  \textbf{SimCLR*} &
  \textbf{ImageNet} &
  \textbf{MoCo v2*} &
  \textbf{SimCLR*} \\
\toprule
\multirow{3}{*}{\textbf{KNN}}          & EuroSAT-L & 78.9 & 76.2 & 79.0 & 69.8 & 76.4 & 75.1 & 72.6 & 79.4 & 81.6 & \textbf{83.4} \\
                                       & BigEarthNet-L & 64.5 & 57.5 & 61.4 & 51.4 & 58.3 & 59.3 & 53.9 & 61.9 & 64.5 & \textbf{67.0} \\
                                       & LC100-L & 56.3 & 56.5 & 56.9 & \textbf{57.7} & 56.6 & 56.2 & 56.8 & 56.5 & 56.0 & 55.8 \\
\midrule
\multirow{3}{*}{\textbf{LP}} & EuroSAT-L & 74.1 & 81.0 & 82.8 & 68.5 & 79.0 & 82.5 & 73.5 & 86.4 & \textbf{88.8} & 77.6 \\
                                       & BigEarthNet-L & 60.9 & 70.2 & 74.0 & 60.3 & 71.9 & 76.1 & 63.1 & 74.9 & \textbf{77.5} & 68.3 \\
                                       & LC100-L & 62.3 & 62.1 & 61.6 & \textbf{63.8} & 60.6 & 60.5 & 62.1 & 60.5 & 59.9 & 63.4 \\
\bottomrule
\end{tabular}%
}
\end{table*}

\begin{description}
    \item[LC100] The LC100 dataset is a multi-label classification dataset which is part of the CopernicusBench~\cite{wang2025towards} benchmark suite. The dataset contains 96 × 96 300m spatial resolution images of Sentinel-3 imagery with 21 multispectral bands. The image classification labels contain 23 categories derived from the 100m land cover product from the Copernicus Global Land Service (CGLS)~\cite{buchhorn2020copernicus}. We utilize the dataset splits provided with the data.

    \item[EuroSAT] The EuroSAT dataset~\cite{helber2019eurosat} is a land cover classification dataset of patches extracted from multispectral Sentinel-2 imagery. The dataset contains 27,000 64 × 64 10m spatial resolution images with 13 bands and labels for 10 land cover categories. We use the dataset splits defined in \cite{neumann2019domain}.

    \item[BigEarthNet] The BigEarthNet dataset \cite{sumbul2019bigearthnet} is a multi-label land cover classification dataset of patches extracted from multispectral Sentinel-2 imagery. The dataset contains 590,326 120 × 120 10m spatial resolution images with 12 multispectral bands and labels for 19 land cover categories. We use the splits provided with the dataset and defined in \cite{sumbul2019bigearthnet}.
\end{description}

\paragraph{Data Acquisition}
We extract the timestamps and geospatial bounds of the imagery of each dataset. We then use this metadata to download corresponding imagery from the Landsat Collection 2 Level-2 dataset using Microsoft's Planetary Computer~\citep{microsoft_open_source_2022_7261897}. We specifically download imagery from the Landsat 8 Operational Land Imager (OLI) and Thermal Infrared Sensor (TIRS)~\citep{engebretson2020olitirs} containing the following 7 multispectral bands -- Coastal/Aerosol, Blue, Green, Red, NIR, SWIR-1, and SWIR-2 -- which correspond to the following band IDs -- \texttt{(SR\_B1, SR\_B2, SR\_B3, SR\_B4, SR\_B5, SR\_B6, SR\_B7)}. All bands are provided with a spatial resolution of 30 m/pixel.

To ensure imagery contains as few invalid pixels as possible, we constrain the extracted Landsat patches to meet a minimum valid pixel ratio of 80\%.

While the BigEarthNet and LC100 imagery have associated timestamps, the EuroSAT dataset does not. Therefore, we initially select a 6-month range on the year of dataset release between $2018/03/01 - 2018/08/31$. For the BigEarthNet and LC100 imagery, we select a range of $\pm{30}$ days from the original timestamp. Due to a few images not being available within the 30-day range, we iteratively increase the range by an additional 30 days until all images are downloaded.

Although the ideal case is to download cloud-free imagery as close to the original timestamp as possible, we observe that a small number of samples require us to lessen the constraints. We initially query for imagery with $\leq 10\%$ cloud cover and with the above timestamp rules. We then sample the image closest to the original timestamp with the least amount of clouds. We log the samples which do not have available imagery and progressively lessen the constraints to $\leq 20\%$ cloud cover and increase the delta of months from the timestamp by 1 until we have collected all remaining samples.

\section{Benchmarks}
\paragraph{Baseline Models}
We evaluate a selection of common convolutional and transformer-based architectures -- \textbf{ResNet-18, ResNet-50, and ViT-S/16}. We compare the performance of model weights pretrained for Landsat imagery vs. a baseline of ImageNet pretraining using the implementations provided in the \texttt{torchgeo} library~\cite{stewart2022torchgeo}. We also include a naive baseline to our experiments, referred to as \textbf{Zonal Stats} where we compute a handcrafted feature embedding for each patch by summarizing each spectral band using per-band mean, standard deviation, minimum, and maximum values and concatenating them into a fixed-length feature vector and applying standard normalization to scale the features to similar ranges. This is analogous to zonal statistics in remote sensing, where aggregate statistics are computed over spatial zones. The following pretraining methods are detailed below:

\begin{table*}[t!]
\centering
\caption{\textbf{Comparison of linear probing results on \texttt{Landsat-Bench} vs. current state-of-the-art on the original datasets.} For EuroSAT we report overall accuracy (OA). For BigEarthNet and LC100 we report micro mean average precision (mAP).}
\label{tab:best_models_landbench}
\resizebox{0.7\textwidth}{!}{%
\begin{tabular}{lcccc}
\toprule
\textbf{Dataset} &
\textbf{Method} &
\textbf{Params (M)} &
\textbf{Metric} &
\textbf{Performance} \\
\toprule

EuroSAT & Copernicus-FM ViT-B/16~\cite{wang2025towards} & 139.5 & OA & 97.9 \\
\rowcolor{Gray}
EuroSAT-L & SSL4EO-L MoCo v2 ViT-S/16~\cite{stewart2023ssl4eo} & 22.1 & OA & 88.8 \\
\midrule

BigEarthNet & SoftCon ViT-B/14~\cite{wang2024multi} & 86.6 & mAP & 86.1 \\
\rowcolor{Gray}
BigEarthNet-L & SSL4EO-L MoCo v2 ViT-S/16~\cite{stewart2023ssl4eo} & 22.1 & mAP & 77.5 \\
\midrule

LC100-S3 & Copernicus-FM ViT-B/16~\cite{wang2025towards} & 139.5 & mAP & 93.3 \\
\rowcolor{Gray}
LC100-L & SSL4EO-L SimCLR ResNet-18~\cite{stewart2023ssl4eo} & 11.7 & mAP & 63.8 \\
\bottomrule
\end{tabular}%
}
\end{table*}

\begin{description}
    \item[ImageNet] We use ImageNet pretrained weights as a baseline, in particular we use the default weights and the first convolutional layer adaptive initialization scheme from the \texttt{timm} library~\cite{rw2019timm} which repeats the RGB weights for additional channels, in our case 7 Landsat multispectral bands.

    \item[SSL4EO-L MoCo v2] We utilize weights from \cite{stewart2023ssl4eo} pretrained using the MoCo v2~\cite{chen2020improved} self-supervised learning method on the SSL4EO-L dataset.

    \item[SSL4EO-L SimCLR] We utilize weights from \cite{stewart2023ssl4eo} pretrained using the SimCLR~\cite{chen2020simple} self-supervised learning method on the SSL4EO-L dataset.
\end{description}

\paragraph{Experiments}
To evaluate existing foundation models' ability to extract useful feature representations for downstream applications, we perform both K-Nearest Neighbor (KNN) (k=5) and Linear Probing (LP) evaluations. To avoid conflating any performance gains with custom implementations of evaluation pipelines, we keep our benchmarks simple and utilize the KNN and Logistic Regression implementations from \texttt{scikit-learn}~\cite{scikit-learn} to operate directly on extracted image embeddings. For Linear Probing of the multi-label datasets (BigEarthNet-L and LC100-L) we employ a One-vs-Rest classifier using a Stochastic Average Gradient (SAG) solver and a maximum of 1,000 iterations. For preprocessing, we use the resize and normalization strategy recommended by \cite{corley2024revisiting}.

\section{Discussion \& Future Work}

\paragraph{Benchmark Results}
The results in Table~\ref{tab:results} provide evidence overall that self-supervised learning methods pretrained on Landsat imagery outperform ImageNet pretraining by a significant margin, e.g. $+4\%$ OA and $+5.1\%$ mAP for the EuroSAT-L and BigEarthNet-L KNN results, respectively. However, selecting which method appears to depend on the task. MoCo v2 tends to significantly outperform SimCLR for linear probing, while the opposite is true for KNN evaluations. For LC100-L, KNN results across all models remained similar; only after linear probing did the differences in model performance become apparent.

\paragraph{Computational Efficiency}
Landsat’s 30m resolution offers a balance between detail and efficiency, avoiding potential increased data and compute costs of 10m sensors like Sentinel-2. For tasks like land-cover classification or change detection, this trade-off can enable faster analysis with minimal loss in performance. As a result, Landsat GFMs can be suitable for resource-constrained geospatial tasks.

For context, assume we want to perform global inference with a ResNet-50 on a NVIDIA V100 GPU with a patch size of $224 \times 224$ at a throughput of 500 images/sec. For the Earth's total land area of 1.49 $\times$ 10\textsuperscript{14} m\textsuperscript{2}, the inference time to process all global patches, assuming non-overlapping patches, using Equation~\ref{eq:inference-time} results in approximately \textbf{1.83 hours} for Landsat-8 and \textbf{16.5 hours} for Sentinel-2 (an increase of \textbf{9}$\mathbf{\times}$ as expected due to the 9$\times$ increase in resolution).

\vspace{-1em}

{\scriptsize
\begin{equation}
\label{eq:inference-time}
\text{T (s)} = \frac{\text{Area}\,(\text{m}^2)}
{\text{Res}^2\,(\text{m\textsuperscript{2}/px\textsuperscript{2}}) \cdot \text{Patch Size}^2\,(\text{px\textsuperscript{2}}) \cdot \text{Throughput}\,(\text{img/s})}
\end{equation}
}

\paragraph{Sensor Ensembling}
Training models for the same tasks for both Landsat and Sentinel constellations can benefit from increased revisit rates. By merging Sentinel's revisit rate of 5 days with Landsat 8 \& 9's rate of 8 days, using an ensemble of multi-sensor models increases the rate at which time-sensitive applications can be performed, such as wildfire~\cite{de2021active} and deforestation~\cite{watch2002global} monitoring.

\paragraph{Thermal Infrared Applications}
While Landsat provides a lower resolution for optical bands than Sentinel-2, its Thermal Infrared Sensor (TIRS) provides 100m thermal surface temperature measurements which are unavailable in the Sentinel-2 constellation. While Sentinel-3 captures thermal infrared, it comes at a 10x lower resolution of 1,000m. This enables applications where thermal monitoring is essential such as urban heat island detection~\cite{keeratikasikorn2018urban} and crop water stress assessment~\cite{ahmad2021review}.

\paragraph{Benchmark Difficulty}
While this can also be seen as a positive, we note there is a significant drop in performance compared to the original datasets -- as seen in Table~\ref{tab:best_models_landbench}. Most notably, LC100-L performance drops from 93.8 to 63.8 mAP for the current best models. This can be attributed to LC100 originally being a Sentinel-3 derived dataset with 21 multispectral bands, potentially containing spectral features for class separability that the Landsat 8 OLI 7 bands do not provide.

\bibliography{refs}
\bibliographystyle{icml2025}

\end{document}